\documentclass[sigconf,nonacm]{acmart}
\renewcommand\footnotetextcopyrightpermission[1]{}

\usepackage{booktabs}
\usepackage{multirow}
\usepackage{graphicx}
\usepackage{amsmath}
\usepackage{xcolor}
\usepackage{colortbl}
\usepackage{subcaption}
\usepackage{microtype}
\usepackage{dblfloatfix}
\usepackage{placeins}
\usepackage{enumitem}

\newcommand{\suite}{CrossView Suite}
\newcommand{\benchtrain}{CrossViewSet}
\newcommand{\bencheval}{CrossViewBench}
\newcommand{\ours}{CrossViewer}
\newcommand{\tokenizer}{ART}
\newcommand{\aligner}{OCVA}

\newcommand{\etal}{\textit{et al.}}
\definecolor{best}{RGB}{180,0,0}
\definecolor{apishade}{RGB}{242,247,255}
\definecolor{openshade}{RGB}{244,249,242}
\definecolor{humanshade}{RGB}{247,244,255}
\definecolor{ourshade}{RGB}{255,244,232}
\definecolor{datasetbadge}{RGB}{144,180,242}
\definecolor{benchbadge}{RGB}{232,190,112}
\newcommand{\best}[1]{\textcolor{best}{\textbf{#1}}}
\newcommand{\yes}{\ensuremath{\checkmark}}
\newcommand{\partialmark}{\ensuremath{\circ}}
\newcommand{\nomark}{--}
\newcommand{\na}{N/A}
\newcommand{\rolebadge}[2]{\begingroup\setlength{\fboxsep}{1.5pt}\colorbox{#1}{\textcolor{white}{\scriptsize\textbf{#2}}}\endgroup}
\newcommand{\dbadge}{\rolebadge{datasetbadge}{D}}
\newcommand{\bbadge}{\rolebadge{benchbadge}{B}}
\newcommand{\bothbadge}{\dbadge\hspace{1pt}\bbadge}

\acmConference[MM '26]{ACM International Conference on Multimedia}{2026}{}

\begin{document}

\title{CrossView Suite: Harnessing Cross-view Spatial Intelligence of MLLMs with Dataset, Model and Benchmark}

\author{Wei Wang}
\affiliation{%
  \institution{Zhejiang University}
  \country{China}}

\author{Yuqian Yuan}
\affiliation{%
  \institution{Zhejiang University}
  \country{China}}

\author{Tianwei Lin}
\affiliation{%
  \institution{Zhejiang University}
  \country{China}}

\author{Wenqiao Zhang}
\authornote{Corresponding author.}
\affiliation{%
  \institution{Zhejiang University}
  \country{China}}

\author{Siliang Tang}
\affiliation{%
  \institution{Zhejiang University}
  \country{China}}

\author{Jun Xiao}
\affiliation{%
  \institution{Zhejiang University}
  \country{China}}

\author{Yueting Zhuang}
\affiliation{%
  \institution{Zhejiang University}
  \country{China}}

\begin{abstract}
Spatial intelligence requires multimodal large language models (MLLMs) to move
beyond single-view perception and reason consistently about objects, visibility,
geometry, and interactions across multiple viewpoints. However, progress in cross-view reasoning remains limited by three major gaps: the scarcity of large-scale well-annotated 
training data, the lack of comprehensive
benchmarks for systematic evaluation, and the absence of explicit alignment
mechanisms that establish object-level consistency across views.
To address these gaps, we thoroughly develop \textbf{\suite} across three
coordinated components: \textbf{\benchtrain{}}, \textbf{\bencheval{}}, and
\textbf{\ours}. Firstly, we introduce a
multi-agent data engine to meticulously curate a large-scale, high-quality cross-view instruction dataset, termed \textbf{\benchtrain{}}, covering  \textbf{17} fine-grained task types with 1.6M samples. 
Second, we meticulously create a scene-disjoint \textbf{\bencheval{}} to comprehensively assess the cross-view spatial understanding capability of an MLLM, evaluating it across various aspects. 
Finally, we propose \textbf{\ours}, a progressive three-stage framework for
cross-view spatial reasoning in MLLMs, following a
\emph{Perception $\rightarrow$ Alignment $\rightarrow$ Reasoning} paradigm.
Our method equips an adaptive spatial region tokenizer to capture fine-grained object representations, and then aligns the multi-view objects explicitly, and thus fuses aligned features for boosting the cross-view inference capacity for MLLMs.
Extensive experiments and analyses show that large-scale training
data, systematic evaluation, and explicit cross-view alignment are all critical
for advancing MLLMs from single-view perception toward real-world spatial intelligence. The project page is available at \url{https://github.com/Thinkirin/Crossview-Suite}.
\end{abstract}



\begin{teaserfigure}
  \centering
  \includegraphics[width=0.98\textwidth]{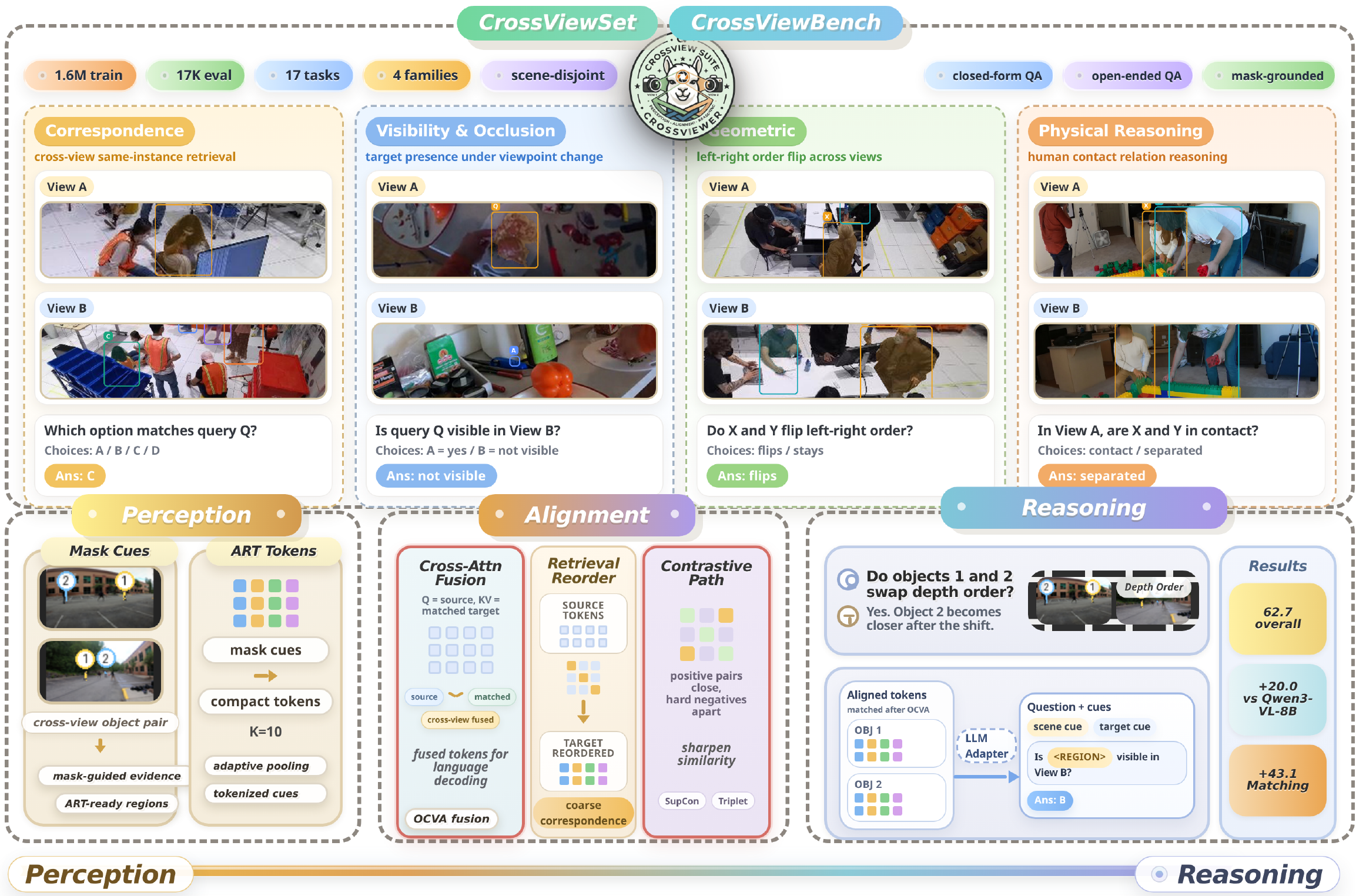}
  \caption{Overview of \ours. \ours{} presents a unified framework for cross-view spatial intelligence in MLLMs, integrating data, benchmark, and model. \benchtrain{} is a large-scale mask-grounded instruction dataset (1.6M samples, 17 tasks), and \bencheval{} is a scene-disjoint benchmark for systematic evaluation. The model follows a progressive Perception--Alignment--Reasoning paradigm, enabling explicit cross-view object alignment and region-guided reasoning.}
  \Description{A dense infographic teaser figure for CrossView Suite.
    The horizontal axis runs from alignment to reasoning, and the vertical axis
    runs from object-level to scene-level understanding. The upper block
    summarizes CrossViewSet and CrossViewBench with example tasks, scale
    statistics, and scene-disjoint evaluation. The lower-left panel shows the
    perception stage with mask cues and ART tokens. The lower-middle panel
    shows the alignment stage with retrieval, supervised contrastive learning,
    and cross-view attention. The lower-right panel shows the reasoning stage
    where aligned region tokens are projected by an LLM adapter into question
    slots together with scene and target cues for answer generation.}
  \label{fig:teaser}
\end{teaserfigure}

\maketitle

\section{Introduction}
\label{sec:intro}

Multimodal large language models (MLLMs) have achieved remarkable progress in
single-view visual understanding, with representative systems such as
Flamingo~\cite{alayrac2022flamingo}, BLIP-2~\cite{li2023blip2},
LLaVA~\cite{liu2024llava}, InternVL~\cite{chen2024internvl}, and
Qwen3-VL~\cite{qwen3vl2025} demonstrating strong performance on captioning and general 
visual question answering. Building on these advances, recent research has begun to extend MLLMs toward richer spatial and 3D reasoning capabilities~\cite{chen2024spatialvlm,cheng2024spatialrgpt,daxberger2025mm,huang2025mllm,zheng2025learning}, which are essential for enabling real-world interaction. 

However, real-world environments are inherently dynamic and multi-perspective: agents must continuously perceive scenes from changing viewpoints rather than relying on a single static image. This shift from static perception to dynamic, multi-view understanding introduces new challenges that cannot be addressed by simply scaling single-view capabilities.
 A model must identify the same physical object across viewpoint changes and use
that aligned representation to support downstream reasoning about visibility,
geometry, and interactions. In real multi-camera settings, judgments such as
cross-view correspondence, occlusion, layout change, and physical contact all
depend on stable object-level consistency across views. This capability is
fundamental for embodied AI, multi-agent collaboration, robotics, and video
analytics, but remains underdeveloped in current MLLMs. We attribute this limitation to the following key challenges.

\textbf{First, The Lack Of Large-Scale Mask-Grounded Cross-View Supervision.}
Existing public
resources such as Ego-Exo4D~\cite{grauman2024egoexo4d}, EgoHumans~\cite{khirodkar2023egohumans}, MessyTable~\cite{cai2020messytable} and
MMPTrack~\cite{han2023mmptrack}
provide synchronized views, identities, or 3D annotations, but they are not
organized as large-scale mask-grounded instruction-tuning data for MLLMs. 
In particular, they do not simultaneously provide reliable and high-quality instance masks for precise object localization and effective background-noise suppression, unified region references, and object-grounded cross-view QA supervision at scale. As a result, while these datasets offer valuable perceptual signals and useful visual cues, they cannot directly provide scalable supervision for teaching MLLMs explicit object-level reasoning across views.

\textbf{Second, The Lack Of A Unified, Multi-Dimensional Benchmark.}
Existing benchmarks, including MMVM, All-Angles Bench, and
MV-ScanQA~\cite{zhou2025mmcorr,yeh2025allanglesbench,mo2025advancing}, reveal
failures in correspondence, viewpoint-sensitive reasoning, and multi-view scene
understanding. Yet each covers only part of the problem, with limited task
breadth, scale, or alignment with real-world mask-grounded multi-camera
reasoning. The field therefore still lacks a unified benchmark that jointly and systematically
evaluates the core capabilities required for multi-view spatial intelligence.

\textbf{Third, The Lack Of An Object-Centric MLLM Framework For Cross-View Reasoning.} Existing region-aware MLLMs can ground
objects within one image, but they do not establish cross-view identity
consistency~\cite{rasheed2024glamm,ma2024groma}. Non-MLLM cross-view methods such as
IDMR~\cite{liu2025idmr}, Wang~\etal~\cite{wang2025towards},
VICI~\cite{zhang2025vici}, and ObjectRelator~\cite{fu2024objectrelator}
have made strong progress on cross-view association. However, these methods are not designed as general-purpose reasoning systems and therefore cannot support the open-ended, object-centric, multi-task reasoning tasks. Current frontier
and recent multi-image MLLMs~\cite{li2024llavanext,wang2024qwen2vl,chen2024internvl,qwen3vl2025,jiang2024mantis} can process multiple images, yet they still rely largely on
implicit fusion inside the language model instead of an explicit object-level
cross-view alignment stage. As a result, they struggle to form stable cross-view object correspondences and capture fine-grained relations among objects across viewpoints.

To address these challenges, we introduce \textbf{\suite}, a unified framework
comprising \benchtrain{}, \bencheval{}, and \ours{}, which together provide an integrated solution for training, evaluation, and deployment.

On the set side, we introduce a
\textbf{multi-agent data engine} to curate \textbf{\benchtrain{}}, a large-scale
cross-view instruction dataset containing \textbf{1.6M} samples from four
public multi-view sources. It spans \textbf{17} fine-grained task types
covering correspondence, visibility and occlusion, geometric reasoning, and
physical reasoning. For sources lacking pixel-level annotations, we generate instance masks and systematically convert raw multi-view observations, object associations, and geometric cues into a unified, object-grounded QA supervision format tailored for MLLMs.

On the evaluation side, we build
\textbf{\bencheval{}}, a comprehensive benchmark with \textbf{17,000} questions
across the same \textbf{17} task types. It evaluates four capability groups:
(i) cross-view correspondence, (ii) visibility and occlusion, (iii) geometric relations, and
(iv) physical reasoning. The benchmark is scene-disjoint from training, uses a
combination of predominantly closed-form QA and complementary open-ended QA
for analysis.

On the model side, we propose \textbf{\ours}, a progressive three-stage
framework for cross-view spatial reasoning in MLLMs. Its design follows three
stages:
\textbf{Perception}, \textbf{Alignment}, and \textbf{Reasoning}.
In the Perception stage, the Adaptive Region Tokenizer (\tokenizer{}) converts each mask-grounded object into compact tokens via adaptive crop-and-scale processing. In the Alignment stage, a retrieval module establishes coarse cross-view correspondences, while the Object-Centric Cross-View Aligner (\aligner{}) integrates cross-attention fusion with contrastive learning to produce consistent object representations across views. In the Reasoning stage, an LLM adapter injects the aligned region tokens at \texttt{<REGION>} positions together with key scene cues, enabling answer generation grounded in explicit object-level evidence.

\begin{table*}[t]
\centering
\caption{Comparison of representative multi-view resources for MLLMs. MV denotes multi-view input, Corr.\ denotes explicit cross-view correspondence supervision, and Scene-disj.\ denotes a scene-disjoint train/eval split. In the Role column, D and B indicate dataset and benchmark, respectively. The symbol $\partialmark$ denotes partial support.}
\label{tab:resource_compare}
\footnotesize
\setlength{\tabcolsep}{4pt}
\renewcommand{\arraystretch}{1.05}
\resizebox{\textwidth}{!}{%
\begin{tabular}{l l c c c c c c c l c}
\toprule
\textbf{Resource} & \textbf{Type} & \textbf{MV} & \textbf{Mask} & \textbf{Corr.} & \textbf{3D/Geo.} & \textbf{QA} & \textbf{Role} & \textbf{Scene-disj.} & \textbf{Scale} & \textbf{Tasks} \\
\midrule
Ego-Exo4D Dataset~\cite{grauman2024egoexo4d} & Raw dataset + training dataset & \yes & \yes & \yes & \yes & \partialmark & \dbadge & \nomark & 5,035 takes / 1,286 h & \na \\
Ego-Exo4D Benchmark~\cite{grauman2024egoexo4d} & Benchmark & \yes & \yes & \yes & \yes & \nomark & \bbadge & \nomark & 3,082 / 842 / 1,121 takes & 8 \\
EgoHumans Dataset~\cite{khirodkar2023egohumans} & Raw dataset & \yes & \partialmark & \yes & \yes & \nomark & \dbadge & \yes & 125K ego RGB / 410K instances & \na \\
EgoHumans Benchmark~\cite{khirodkar2023egohumans} & Benchmark & \yes & \partialmark & \yes & \yes & \nomark & \bbadge & \yes & 77,260 / 47,740 images & 4 \\
MessyTable~\cite{cai2020messytable} & Raw dataset + benchmark & \yes & \nomark & \yes & \yes & \nomark & \bothbadge & \nomark & 5,579 scenes / 50,211 images & 1 \\
MMPTrack~\cite{han2023mmptrack} & Raw dataset + benchmark & \yes & \nomark & \yes & \yes & \nomark & \bothbadge & \nomark & 261 / 153 / 162 min ($\sim$2.98M GT) & 2 \\
MMVM SFT Dataset~\cite{zhou2025mmcorr} & Training dataset & \yes & \yes & \yes & \nomark & \yes & \dbadge & \nomark & 220K QA samples & \na \\
MMVM Benchmark~\cite{zhou2025mmcorr} & Benchmark & \yes & \yes & \yes & \nomark & \yes & \bbadge & \nomark & 1,510 examples & 1 \\
All-Angles Bench~\cite{yeh2025allanglesbench} & Benchmark & \yes & \nomark & \yes & \nomark & \yes & \bbadge & \nomark & 2,132 examples & 6 \\
MV-ScanQA~\cite{mo2025advancing} & Benchmark & \yes & \nomark & \yes & \yes & \yes & \bbadge & \yes & 10K examples & 1 \\
\midrule
\rowcolor{ourshade}
\textbf{\benchtrain{} (ours)} & Training dataset & \yes & \yes & \yes & \yes & \yes & \dbadge & \yes & 1.643M samples & 17 \\
\rowcolor{ourshade} \textbf{\bencheval{} (ours)} & Benchmark & \yes & \yes & \yes & \yes & \yes & \bbadge & \yes & 17K questions & 17 \\
\bottomrule
\end{tabular}%
}
\end{table*}

This unified setup enables end-to-end investigation of explicit cross-view alignment.
\ours{} reaches \textbf{62.7\%} overall accuracy on \bencheval{}, outperforming
Qwen3-VL-8B by \textbf{20.0\%}, and reaches 49.5\% accuracy on valid
MMVMBench cases. These results highlight the importance of scalable
training data, scene-disjoint evaluation, and explicit cross-view alignment
for advancing MLLMs toward multi-view spatial intelligence. Our main
contributions are summarized as follows:

\begin{itemize}[leftmargin=2em]
    \item \textbf{We build \benchtrain{}}, a 1.6M-sample cross-view instruction dataset with mask grounding and object-level QA supervision across 17 tasks.

    \item \textbf{We introduce \bencheval{}}, a comprehensive benchmark with 17K questions  covering correspondence, occlusion, geometry, and physical reasoning in multi-view settings.

    \item \textbf{We propose \ours{}}, a progressive three-stage framework for cross-view spatial reasoning that enables object-level reasoning across multiple views through explicit alignment.

\end{itemize}

\section{Related Work}

\subsection{Multimodal Large Language Models}
MLLMs have evolved from early adapter-style systems such as
Flamingo~\cite{alayrac2022flamingo} and BLIP-2~\cite{li2023blip2} to
instruction-tuned models including InstructBLIP~\cite{dai2023instructblip},
LLaVA~\cite{liu2024llava}, Qwen-VL~\cite{bai2023qwenvl}, and recent successors
such as LLaVA-NeXT~\cite{li2024llavanext},
Qwen2-VL~\cite{wang2024qwen2vl}, InternVL~\cite{chen2024internvl}, and
Qwen3-VL~\cite{qwen3vl2025}. These models improve perception, multi-image
input, and open-ended reasoning, but still rely mainly on implicit fusion
rather than explicit object-level cross-view alignment.
Recent specialized and modular MLLMs further explore domain adaptation and
dynamic expert tuning, as exemplified by medical VLM adaptation and dynamic
visual-language expert routing~\cite{lin2025healthgpt,zhang2024hyperllava}.

Related efforts extend MLLMs toward spatial and 3D reasoning, including
SpatialVLM~\cite{chen2024spatialvlm},
SpatialRGPT~\cite{cheng2024spatialrgpt}, MM-Spatial~\cite{daxberger2025mm},
MLLM-for3D~\cite{huang2025mllm}, and
Video-3D LLM~\cite{zheng2025video}. However, these methods emphasize geometry
and 3D scene understanding more than synchronized instance-level alignment
across views. Empirical studies such as
Eyes~Wide~Shut~\cite{tong2024eyeswideshut},
MMVM~\cite{zhou2025mmcorr}, and
All-Angles~Bench~\cite{yeh2025allanglesbench} further expose weak cross-view
consistency, motivating the explicit alignment design in \ours.

\subsection{Fine-Grained Visual Understanding}
Promptable segmentation models such as SAM~\cite{kirillov2023sam},
SAM~2~\cite{ravi2024sam2}, and LISA~\cite{lai2024lisa} provide strong mask
generation and language-guided segmentation. Region-aware MLLMs such as
Ferret~\cite{you2024ferret}, Osprey~\cite{yuan2024osprey},
GLaMM~\cite{rasheed2024glamm}, RegionGPT~\cite{guo2024regiongpt}, and
AnyRef~\cite{he2024anyref} extend instruction tuning to arbitrary regions and
fine-grained visual references. Recent object-centric and video referring
systems further broaden this direction toward spatio-temporal grounding,
segmentation, editing, and generation~\cite{yuan2025videorefer,yuan2026lmms,zhong2026unified}.
PixelRefer~\cite{yuan2025pixelrefer} is especially related because it
compresses mask regions into compact tokens, yet these methods remain
fundamentally single-view. \ours~extends this mask-grounded setting to
explicit cross-view correspondence and multi-view spatial reasoning.

\subsection{Cross-View Alignment and Correspondence}
Classical methods such as SuperGlue~\cite{sarlin2020superglue},
LoFTR~\cite{sun2021loftr}, and diffusion-based correspondence
analysis~\cite{tang2023emergent} focus on low-level point or patch matching.
More semantic vision-language work such as
MMVM~\cite{zhou2025mmcorr}, IDMR~\cite{liu2025idmr},
VICI~\cite{zhang2025vici}, and ObjectRelator~\cite{fu2024objectrelator}
studies paired-image correspondence, retrieval, localization, or ego/exo
relations, but still targets narrower matching or reasoning settings rather
than combining large-scale mask-grounded training, explicit object-token
alignment, and unified multi-task QA evaluation.

\subsection{Multi-View Datasets and Benchmarks}
Representative multi-view resources include Ego-Exo4D, EgoHumans,
MessyTable, and
MMPTrack~\cite{grauman2024egoexo4d,khirodkar2023egohumans,cai2020messytable,han2023mmptrack}.
Benchmark resources such as MMVM, All-Angles Bench, and
MV-ScanQA~\cite{zhou2025mmcorr,yeh2025allanglesbench,mo2025advancing} evaluate
narrower slices of cross-view reasoning (Table~\ref{tab:resource_compare}).
Egocentric object benchmarks such as EOC-Bench~\cite{yuan2025eoc} also
highlight the importance of identifying, recalling, and forecasting objects in
dynamic first-person scenes.
These resources provide synchronized views, identities, or 3D signals, but not
a large-scale mask-grounded MLLM corpus with unified QA. By
contrast, \benchtrain{} and \bencheval{} couple large-scale training and
scene-disjoint evaluation under the same taxonomy.


\section{\benchtrain{} and \bencheval{}}
\label{sec:benchmark}
In this section, we introduce \benchtrain{}, a large-scale dataset designed to equip models with cross-view reasoning capability, and \bencheval{}, a comprehensive benchmark for systematic evaluation and fair comparison.

\subsection{Source Data Collection}
Our dataset and benchmark integrate four established multi-view resources:
Ego-Exo4D~\cite{grauman2024egoexo4d} for synchronized ego/exo activity videos,
EgoHumans~\cite{khirodkar2023egohumans} for dynamic multi-person scenes,
MMPTrack~\cite{han2023mmptrack} for surveillance-style multi-camera
association, and MessyTable~\cite{cai2020messytable} for cluttered tabletop
instance association. Together, they cover human activities, multi-person
interaction, surveillance, and tabletop-object scenarios across mixed ego/exo
viewpoints, static and environments, and both human-centered and
object-centered targets.

\subsection{Question Taxonomy}
We introduce a hierarchical taxonomy to systematically characterize cross-view spatial reasoning capabilities. The taxonomy comprises 17 fine-grained task types (Q1--Q17), organized into four categories, as summarized in Table~\ref{tab:qtypes}.

\textbf{Correspondence} (Q1--Q5) focuses on identifying and matching the same physical instance across different viewpoints. \textbf{Visibility \& Occlusion} (Q6--Q8) evaluates cross-view presence and reasoning about occlusions and occluders. \textbf{Geometric} (Q9--Q16) encompasses tasks such as nearest-neighbor retrieval, scale estimation, completeness assessment, arrangement inversion, and displacement reasoning. Finally, \textbf{Physical Reasoning} (Q17) examines whether visual overlap implies physical contact.
All questions are formulated as multiple-choice (four options) or binary where appropriate. Multiple-choice items include distractors where needed.

\begin{table}[!t]
\centering
\caption{Question taxonomy of \benchtrain{} and \bencheval{}.}
\label{tab:qtypes}
\small
\setlength{\tabcolsep}{4pt}
\renewcommand{\arraystretch}{1.05}
\begin{tabular}{c p{4.55cm} r r}
\toprule
\textbf{ID} & \textbf{Task} & \textbf{Set} & \textbf{Bench} \\
\midrule
\rowcolor{openshade}
\multicolumn{2}{l}{\textbf{Correspondence}} & \textbf{632K} & \textbf{5K} \\
Q1  & Single-object cross-view retrieval & 99K & 1K \\
Q2  & Two-candidate retrieval & 107K & 1K \\
Q3  & One-of-three identification & 191K & 1K \\
Q4  & Multi-object group matching & 119K & 1K \\
Q5  & Binary instance verification & 116K & 1K \\
\rowcolor{apishade}
\multicolumn{2}{l}{\textbf{Visibility \& Occlusion}} & \textbf{260K} & \textbf{3K} \\
Q6  & Cross-view visibility detection & 116K & 1K \\
Q7  & Exclusive-presence identification & 119K & 1K \\
Q8  & Occluder source localization & 25K & 1K \\
\rowcolor{openshade}
\multicolumn{2}{l}{\textbf{Geometric}} & \textbf{743K} & \textbf{8K} \\
Q9  & Cross-view nearest-neighbour retrieval & 36K & 1K \\
Q10 & Scale change detection & 109K & 1K \\
Q11 & Single-view completeness comparison & 22K & 1K \\
Q12 & Cross-view completeness comparison & 100K & 1K \\
Q13 & Left-right arrangement flip detection & 121K & 1K \\
Q14 & Cross-view displacement comparison & 117K & 1K \\
Q15 & Depth-relation consistency & 116K & 1K \\
Q16 & Relative scale-growth comparison & 122K & 1K \\
\rowcolor{apishade}
\multicolumn{2}{l}{\textbf{Physical Reasoning}} & \textbf{8K} & \textbf{1K} \\
Q17 & Visual overlap vs.\ physical contact & 8K & 1K \\
\midrule
\textbf{Total} &  & \textbf{1,643K} & \textbf{17K} \\
\bottomrule
\end{tabular}
\end{table}
Although \bencheval{} primarily adopts closed-form QA for stable and objective
large-scale evaluation, it also includes a small amount of open-ended
questions for supplementary analysis. These samples remove answer options while
preserving the same mask grounding, allowing us to test cross-view reasoning
beyond predefined choices without changing the main evaluation protocol.

\subsection{Automatic Generation Pipeline}
We formulate the construction of \benchtrain{} and \bencheval{} as a
tool-augmented multi-agent engine built on public multi-view sources and their
original annotations. A coordinator agent calls tools and sub-agents for
perception completion, cross-view association, question generation,
open-ended rewriting, quality control, and final human verification.
This design is also motivated by prior data-centric learning pipelines that
combine pseudo labeling, active annotation, partial-label learning, and
uncertainty-aware transfer to reduce annotation cost while controlling label
noise~\cite{zhang2022boostmis,zhang2023learning,zhang2024revisiting}.

The pipeline has five stages: (1) the coordinator agent calls tools such as
SAM~2~\cite{ravi2024sam2} to recover missing masks from boxes or keypoints and
filter low-confidence outputs; (2) it derives geometric supervision and
structured cross-view cues from 3D coordinates and camera extrinsics; (3) it
instantiates rule-based QA templates, yielding over \textbf{8M} candidates
before filtering; (4) for the supplementary open-ended setting, it first calls
a rewriting sub-agent to convert selected closed-form samples into grounded
short-answer questions, and then a verification sub-agent to remove ambiguous
or low-quality cases; and (5) after automatic generation, filtering,
rewriting, and consistency checks, the pipeline retains about \textbf{1.6M}
training samples and \textbf{17,000} evaluation questions, followed by human
re-inspection on the evaluation set and a stratified training subset. In this
reviewed portion, \textbf{98.2\%} of training samples and \textbf{96.8\%} of
evaluation samples are kept directly, while the remainder are corrected or
removed.

\begin{figure}[t]
  \centering
  \includegraphics[width=\linewidth]{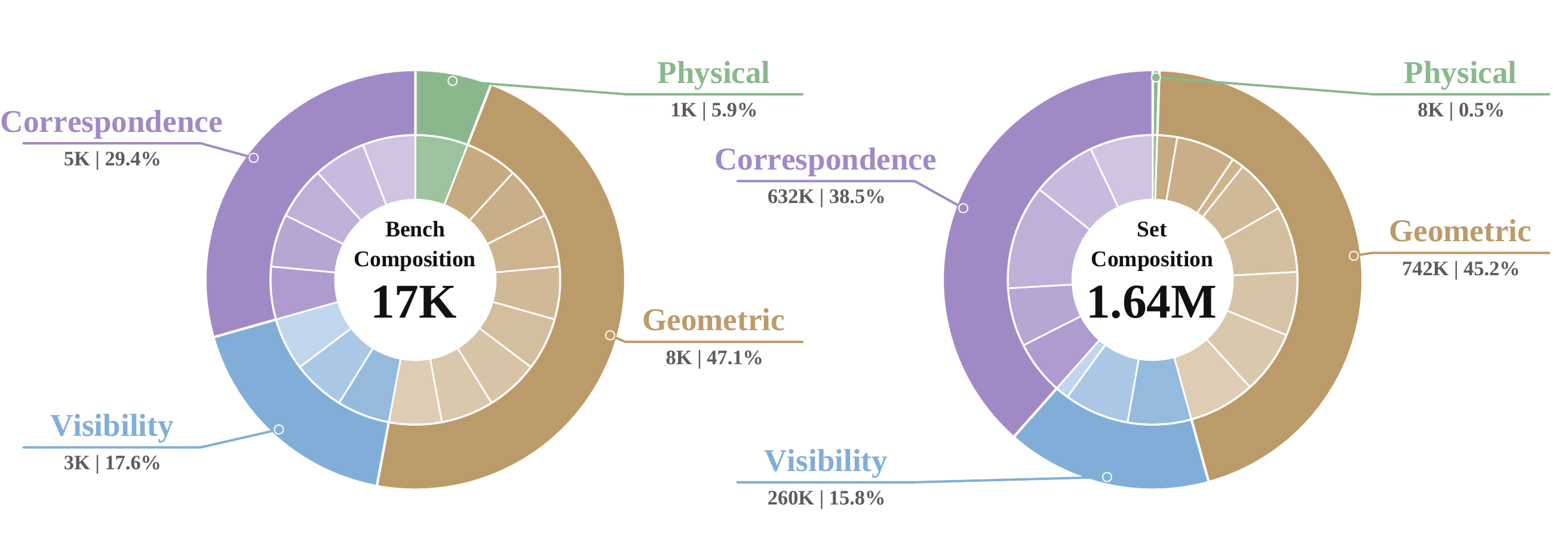}
  \caption{Source and task composition of \benchtrain{} and \bencheval{}.}
  \label{fig:vis_data}
\end{figure}

\subsection{Data Characteristics}
\benchtrain{} and \bencheval{} provide a large-scale, scene-disjoint testbed
for cross-view spatial reasoning with \textbf{1.6M} training samples,
\textbf{17,000} evaluation questions, and \textbf{17} task types. All samples
are region-grounded and carry consistent cross-view identities, enabling
supervision of object correspondence and relations.



\section{Method}
\label{sec:method}

\begin{figure*}[t]
  \centering
  \includegraphics[width=0.75\textwidth]{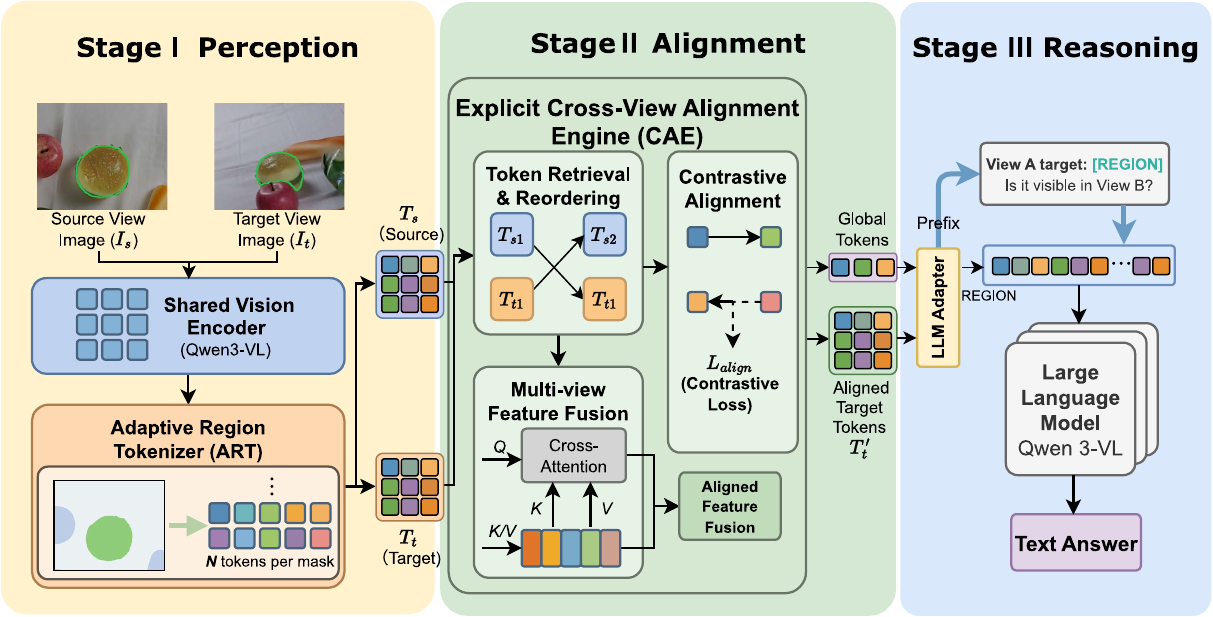}
  \caption{Overview of \ours. Stage I extracts mask-grounded object tokens, Stage II aligns them across views, and Stage III performs region-guided reasoning.}
  \label{fig:arch}
\end{figure*}

\subsection{Problem Formulation}
\label{sec:formulation}

Given $N$ images $\mathcal{I}=\{I^{(n)}\}_{n=1}^{N}$ with associated mask sets
$\mathcal{M}^{(n)}=\{m_1^{(n)},\ldots,m_{K_n}^{(n)}\}$, where
$I^{(n)}\!\in\!\mathbb{R}^{H\times W\times 3}$ and
$m_i^{(n)}\!\in\!\{0,1\}^{H_0\times W_0}$, the model learns
\begin{equation}
  P\!\left(a\;\middle|\;\mathcal{I},\,\mathcal{M},\,q;\,\Theta\right),
  \label{eq:prob}
\end{equation}
for a question $q$, where $a$ is the text answer and $\Theta$ denotes trainable
parameters.

Unlike single-view VQA, the model must establish correspondences across
viewpoints. Let $\mathcal{C}\subseteq\{(i,n,j,n'): n\neq n'\}$ denote the
ground-truth cross-view correspondence set, where $(i,n,j,n')$ indicates that
object $i$ in view $n$ and object $j$ in view $n'$ refer to the same physical
entity; the goal is to \emph{align} co-referent objects in feature space and
\emph{reason} over the aligned representations to answer the question.
\subsection{Model Overview}
\label{sec:overview}
As shown in Fig.~\ref{fig:arch}, 
\ours~follows three conceptual stages: Perception, Alignment and Reasoning. In
the perception stage, a frozen Qwen3-VL-8B vision encoder extracts per-view
feature maps $\mathbf{F}^{(n)}\in\mathbb{R}^{H_v\times W_v\times D_v}$, and
Adaptive Region Tokenizer (\tokenizer{}) converts each mask-grounded object
into up to $K$ compact tokens $\mathbf{T}_i^{(n)}\in\mathbb{R}^{K\times D_v}$.
A retrieval module then computes coarse cross-view correspondences and reorders
the target-view tokens into $\tilde{\mathbf{T}}^{(b)}$. In the alignment
stage, Object-Centric Cross-View Aligner (\aligner{}) explicitly fuses each
source object with its matched target counterpart while producing contrastive
embeddings. Finally, in the reasoning stage, an LLM adapter projects aligned
visual features into the language space and injects them at
\texttt{<REGION>} positions together with global and target-object cues for
answer generation.

\subsection{1st Stage: Fine-Grained Perception}
\label{sec:perception}

\noindent\textbf{Adaptive Region Tokenizer.} 
\label{sec:saot}
Objects can vary substantially in scale and visible extent across viewpoints,
making fixed-resolution region pooling inefficient and unstable for cross-view
matching. We therefore propose an Adaptive Region Tokenizer (\tokenizer{}) to convert each
mask-grounded object into a compact token sequence. Given feature map
$\mathbf{F}^{(n)}$ and mask $m$, let $\mathcal{B}=(t,l,b,r)$ be the tight
bounding box of the mask with $2p$ padding, where $p$ is the patch size. To
retain roughly $K$ informative tokens after pooling, we compute a scale factor
from the mask area $|m| = \sum_{x,y} m_{x,y}$:
\begin{equation}
  s = \left\lceil\sqrt{\frac{K \cdot P^2}{|m|}}\right\rceil,\quad
  h_r = \left\lceil\frac{(b-t)\cdot s}{P}\right\rceil\cdot P,\quad
  w_r = \left\lceil\frac{(r-l)\cdot s}{P}\right\rceil\cdot P,
  \label{eq:scale}
\end{equation}
where $P=14$ is the patch size and $K=10$ is the target token count.
The crop is bilinearly resized to $(h_r/P)\times(w_r/P)$ patch tokens so that
small objects are upsampled to maintain token density while large objects are
only mildly downsampled to control cost.

We then add learnable absolute position embeddings, retain only masked tokens,
run lightweight k-means, and project the resulting centroids with a shared
two-layer MLP to obtain $\mathbf{T}_i^{(n)}$. Object token sequences are padded
to a uniform length with token-validity masks. This crop-first design reduces
background interference while preserving efficient batched multi-view
processing.

\noindent\textbf{Cross-View Object Retrieval.} 
\label{sec:retrieval}
Exhaustively interacting every object across views would introduce many
irrelevant pairs. Before feature fusion, the model must therefore determine
which object in view $b$ corresponds to object $i$ in view $a$. We masked-mean-pool the valid
(non-padding) \tokenizer{} tokens of each object, where
$\mathcal{V}_i^{(a)}$ denotes the valid token positions of object $i$, and
project them through a shared contrastive head $g(\cdot)$:
\begin{align}
  \bar{\mathbf{t}}_i^{(a)}
  &= \frac{1}{|\mathcal{V}_i^{(a)}|}\sum_{k\in\mathcal{V}_i^{(a)}} T_{i,k}^{(a)},
  \nonumber\\
  \mathbf{z}_i^{(a)}
  &= \frac{g\!\left(\bar{\mathbf{t}}_i^{(a)}\right)}
           {\left\|g\!\left(\bar{\mathbf{t}}_i^{(a)}\right)\right\|_2},
  \qquad
  S_{ij} = {\mathbf{z}_i^{(a)}}^{\!\top}\mathbf{z}_j^{(b)},
  \label{eq:simmat}
\end{align}
where $\mathbf{z}_i^{(a)}\in\mathbb{S}^{d_c-1}$, $\mathbf{S}=[S_{ij}]$, and
$g(\cdot)$ is shared with \aligner{}; we set $d_c{=}256$.

\ours~supports three matching modes. GT follows the dataset's canonical
ordering and is used only as an oracle analysis branch. Greedy selects the
highest-scoring unmatched pairs, while Hungarian solves the global 1-to-1
assignment when extra latency is acceptable. The recovered mapping $\pi$
reorders target-view tokens into $\tilde{\mathbf{T}}^{(b)}$ before
\aligner{}. For correspondence-heavy questions (Q1--Q5), $\mathbf{S}$ provides
the explicit cross-view signal, while final answers still come from decoder
scoring.

\subsection{2nd Stage: Cross-View Alignment}
\label{sec:cvam}

Retrieval provides only a coarse correspondence hypothesis and does not yet
model token-level interaction between matched objects. After token reordering,
\aligner{} processes each source-view object token
sequence $\mathbf{T}_i^{(a)}$ together with its matched target-view counterpart
$\tilde{\mathbf{T}}_i^{(b)}$ to build a unified cross-view representation. It
has two parallel paths: a cross-attention fusion path, which injects
target-view context into source-view tokens for language decoding, and a
contrastive alignment path, which trains the shared projection head
$g(\cdot)$ to sharpen the similarity matrix $\mathbf{S}$
(Eq.~\eqref{eq:simmat}).

\noindent\textbf{Cross-Attention Fusion Path.} 
Because each view often reveals only partial evidence of the same object, the
source tokens should explicitly absorb complementary cues from the matched
target view. Using $\mathbf{T}_i^{(a)}$ as queries and $\tilde{\mathbf{T}}_i^{(b)}$ as keys
and values, we apply multi-head cross-attention:
\begin{equation}
  \hat{\mathbf{T}}_i^{(a)}
  = \mathrm{LN}\!\left(
      \mathbf{T}_i^{(a)}
      + \mathrm{MHA}\!\left(
          \mathbf{T}_i^{(a)},\,
          \tilde{\mathbf{T}}_i^{(b)},\,
          \tilde{\mathbf{T}}_i^{(b)}
        \right)
    \right),
  \label{eq:mha}
\end{equation}
followed by a standard FFN with residual layer normalisation.

\noindent\textbf{Contrastive Alignment Path.} 
Fusion contextualizes features for decoding, but it does not explicitly enforce
stable identity structure in the embedding space. The contrastive path uses the
shared projection head $g(\cdot)$ to place
co-referent objects close on the unit hypersphere and separate non-matches.
Object tokens $\mathbf{T}_i^{(a)}$ and $\tilde{\mathbf{T}}_i^{(b)}$ are
independently mean-pooled and projected through $g(\cdot)$ to obtain
embeddings $\mathbf{z}_i^{(a)},\mathbf{z}_i^{(b)}\in\mathbb{S}^{d_c-1}$
(Eq.~\eqref{eq:simmat}).

\noindent\textbf{Supervised Contrastive Loss.}
For an aligned view pair, let $M$ denote the number of valid non-global object
pairs after masking. We concatenate all valid object embeddings from both views
into $2M$ vectors with track-id labels and optimize a supervised contrastive
loss $\mathcal{L}_{\mathrm{SC}}$ following~\cite{DBLP:journals/corr/abs-2004-11362}.
With temperature $\tau{=}0.07$, embeddings sharing the same track-id are
treated as positives, so $\mathcal{L}_{\mathrm{SC}}$ can exploit multiple
positives for the same physical entity within a batch.

\noindent\textbf{Online Hard-Negative Triplet Loss.}
We additionally use an online hard-negative triplet term, where the negative
for each anchor is mined from the valid target-side embeddings in the current
mini-batch. The resulting triplet loss $\mathcal{L}_{\mathrm{tri}}$ uses the
matched object as the positive, the highest-similarity non-match
$\mathbf{z}_{j_i^*}^{(b)}$ selected from that set as the negative, Euclidean distance
$d(\cdot,\cdot)$, and margin $m{=}0.5$. This auxiliary term tightens the
decision boundary around hard confusions.

\subsection{3rd Stage: Region-Guided Reasoning}
\label{sec:reasoning}

Even after alignment, object-level evidence can be diluted if it is passed to
the decoder only as generic multi-image features. After explicit alignment,
\ours{} performs answer generation with a standard
LLM decoder, but now conditioned on aligned object evidence rather than raw
multi-image fusion alone. Each question contains one or more
\texttt{<region>} placeholders together with a \texttt{region\_refs} list of
$(v,\,i)$ tuples indicating the referenced objects. For each referenced object,
the aligned token sequence $\hat{\mathbf{T}}_i^{(v)}$ is projected by the LLM
adapter $h(\cdot)$ into region embeddings
$\mathbf{R}_i^{(v)} = h\!\left(\hat{\mathbf{T}}_i^{(v)}\right)$, and each
placeholder expands into $c_i$ \texttt{<REGION>} tokens whose embeddings are
replaced by $\mathbf{R}_i^{(v)}$.

Beyond region-token injection, masked-pooled aligned features produce a global
scene token and a target-object summary token, which are prepended as visual
prefix cues before the question tokens. Let $\mathbf{g}$ denote the projected
global feature and $\mathbf{o}$ the projected target-object feature. The final
reasoning input is formed as
\begin{equation}
  \mathbf{e}_q =
  \big[\mathbf{g};\,\mathbf{o};\,\mathrm{Tok}(q,\{\texttt{<REGION>}\!\mapsto\!\mathbf{R}\})\big],
  \label{eq:reasoning_input}
\end{equation}
where $\mathrm{Tok}(\cdot)$ denotes token embeddings after region replacement.
The LLM then autoregressively predicts the textual answer from
$\mathbf{e}_q$. Although the decoder itself is standard, its reasoning is
grounded on explicitly aligned cross-view object representations, allowing
correspondence, visibility, geometric, and physical questions to be answered
through a unified interface.

\noindent\textbf{Training Objective.}
Answer supervision alone is insufficient to preserve stable cross-view
correspondences during optimization. \ours~is trained with standard
autoregressive answer cross-entropy together with the two cross-view alignment
losses above:
\begin{equation}
  \mathcal{L}
  = \lambda_{\mathrm{VQA}}\,\mathcal{L}_{\mathrm{VQA}}
  + \lambda_{\mathrm{SC}}\,\mathcal{L}_{\mathrm{SC}}
  + \lambda_{\mathrm{tri}}\,\mathcal{L}_{\mathrm{tri}},
  \label{eq:total}
\end{equation}
where $\mathcal{L}_{\mathrm{VQA}}$ is the token-level answer cross-entropy on
the gold response, and $\mathcal{L}_{\mathrm{SC}}$ and
$\mathcal{L}_{\mathrm{tri}}$ are the two alignment losses defined above. In all
experiments, we set $\lambda_{\mathrm{VQA}}{=}0.5$,
$\lambda_{\mathrm{SC}}{=}1.0$, and $\lambda_{\mathrm{tri}}{=}0.1$.

\begin{figure*}[!t]
  \centering
  \hspace*{-0.045\textwidth}%
  \begin{minipage}[t]{0.39\textwidth}
    \vspace{0pt}
    \centering
    \includegraphics[height=0.176\textheight]{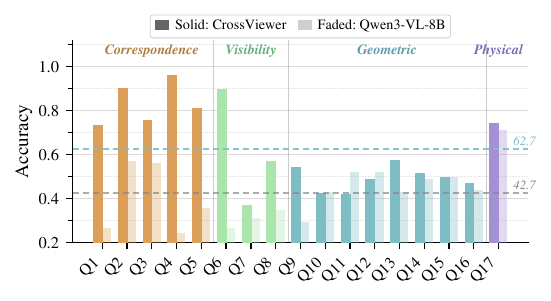}
    \par\vspace{2pt}
    {\footnotesize\textbf{(a)} Per-type accuracy on \bencheval{}.\par}
  \end{minipage}\hspace{0.004\textwidth}%
  \begin{minipage}[t]{0.27\textwidth}
    \vspace{0pt}
    \centering
    \includegraphics[height=0.176\textheight]{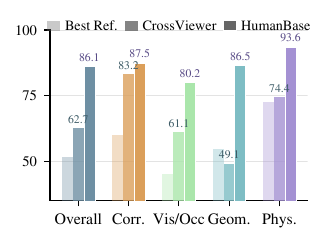}
    \par\vspace{2pt}
    {\footnotesize\textbf{(b)} Gap to the strongest reference and HumanBase.\par}
  \end{minipage}\hspace{0.004\textwidth}%
  \begin{minipage}[t]{0.32\textwidth}
    \vspace{0pt}
    \centering
    \includegraphics[height=0.176\textheight]{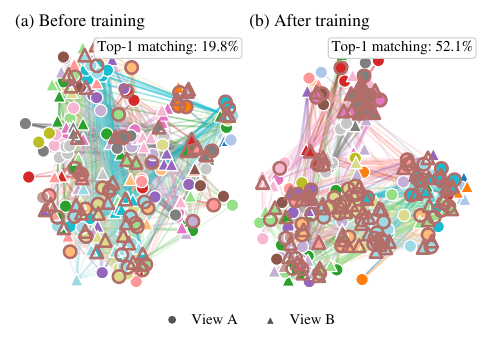}
    \par\vspace{2pt}
    {\footnotesize\textbf{(c)} t-SNE of Q1 correspondence embeddings before and after training.\par}
  \end{minipage}
  \caption{Overview panels of \suite{} and \ours{}: per-type gains, gap to HumanBase, and t-SNE of Q1 correspondence embeddings before and after training.}
  \label{fig:small_panels_top}
\end{figure*}

\section{Experiments}
\label{sec:exp}

\subsection{Implementation Details}
We instantiate \ours~on top of Qwen3-VL-8B-Instruct~\cite{qwen3vl2025}.
Unless otherwise noted, \tokenizer{} uses $K=10$ object tokens, the Object-Centric Cross-View
Aligner adopts 8 attention heads, and the contrastive embedding dimension is set
to $d_{\text{contrast}}=256$.
Training mainly targets the lightweight modules introduced by \ours, including
the \tokenizer{} projection layers, the cross-view alignment blocks, and the
language-side adaptation layers.
We use AdamW with $\text{lr}=10^{-4}$ and a cosine decay schedule on
4$\times$A100 GPUs.
In the default configuration, we use greedy cross-view matching, fused
region-token injection, and pairwise aggregation for VQA over the available
view pairs; contrastive supervision follows the paired-view training path set by
the configuration. Unless otherwise specified, training uses the full
\benchtrain{} split produced by our data generation pipeline.

\begin{table}[t]
\centering
\caption{Performance comparisons on \bencheval{}. The best results are \textbf{bold} and the second-best results are \underline{underlined}.}
\label{tab:main}
\setlength{\tabcolsep}{4pt}
\renewcommand{\arraystretch}{1.06}
\newcommand{\tabbest}[1]{\textbf{#1}}
\newcommand{\tabsecond}[1]{\underline{#1}}
\resizebox{\columnwidth}{!}{%
\begin{tabular}{lccccc}
\toprule
Method & Overall & Corr. & Vis/Occ & Geometric & Physical \\
\midrule
\rowcolor{humanshade}
	HumanBase                         & 86.1 & 87.5 & 80.2 & 86.5 & 93.6 \\
	\midrule
\rowcolor{apishade}
\multicolumn{6}{l}{\textbf{Proprietary Models}} \\
Gemini-3.1-Pro                    & 51.5 & \tabsecond{60.0} & 39.0 & 50.5 & 56.0 \\
GPT-5.2                           & 49.5 & 41.5 & \tabsecond{45.1} & \tabbest{54.5} & 58.3 \\
Grok-4-Fast                       & 42.5 & 38.8 & 33.3 & 45.8 & 59.0 \\
\rowcolor{openshade}
\multicolumn{6}{l}{\textbf{Open-source Models}} \\
Qwen3.5-397B            & \tabsecond{51.7} & 50.1 & 41.0 & \tabsecond{54.1} & \tabsecond{72.6} \\
Qwen3.5-35B                       & 50.1 & 48.3 & 39.4 & 53.3 & 65.6 \\
Qwen3-VL-235B-A22B-Instruct       & 47.7 & 46.3 & 37.5 & 50.2 & 65.3 \\
InternVL2.5-38B                   & 45.9 & 43.8 & 36.2 & 48.4 & 65.5 \\
Qwen2.5-VL-72B                    & 45.2 & 43.2 & 35.6 & 47.8 & 63.2 \\
LLaVA-Video-Qwen2-72B             & 44.2 & 42.6 & 34.9 & 46.7 & 60.1 \\
InternVL2.5-78B                   & 43.7 & 42.4 & 34.3 & 46.4 & 56.8 \\
LLaVA-OneVision-Qwen2-72B         & 43.6 & 42.2 & 34.5 & 45.7 & 61.1 \\
DeepSeek-VL2                      & 42.8 & 41.6 & 33.8 & 44.8 & 59.8 \\
Qwen3-VL-8B                       & 42.7 & 40.1 & 30.7 & 45.3 & 71.1 \\
	InternVL2.5-4B                    & 42.0 & 40.2 & 33.5 & 44.4 & 57.3 \\
	InternVL2.5-2B                    & 40.0 & 38.8 & 31.9 & 42.2 & 52.7 \\
	\midrule
	\rowcolor{ourshade}
	\textbf{\ours~(ours)} & \tabbest{62.7} & \tabbest{83.2} & \tabbest{61.1} & 49.1 & \tabbest{74.4} \\
\bottomrule
\end{tabular}%
}
\end{table}

\subsection{Evaluation Setup}
We evaluate \ours{} on both \bencheval{} and MMVMBench~\cite{zhou2025mmcorr}. On \bencheval{}, we report accuracy for each question type and overall accuracy. For \ours, referenced objects are provided through the native \texttt{<REGION>} token-replacement interface described in Sec.~\ref{sec:reasoning}. For baseline MLLMs without region-token inputs, we instead render each referenced mask as a semi-transparent colored overlay and rewrite the question to use the corresponding color cue, yielding a unified object-grounded evaluation protocol across models.


\subsection{Main Results}

\noindent\textbf{\bencheval{}}. 
Table~\ref{tab:main} compares \ours~with 15 reference MLLMs spanning both
proprietary and open-source models, while 
Fig.~\ref{fig:small_panels_top}(b) highlights the remaining gap to HumanBase.
\ours achieves the best overall performance at 62.7 accuracy, surpassing the
strongest reference model, Qwen3.5-397B, by 11.0 points and the same-backbone
Qwen3-VL-8B baseline by 20.0 points. The latter gap is particularly important:
it shows that the gains are not explained by model scale or broader pretraining
alone, but by the explicit object-centric alignment and reasoning design of
\ours.

Fig.~\ref{fig:small_panels_top} (a) shows that the gains are broad
rather than concentrated on a single task family. The largest improvements
appear on Correspondence and Visibility/Occlusion, where general-purpose MLLMs
still rely mostly on implicit multi-image matching and therefore struggle under
viewpoint change and partial observation. Larger frontier models remain more
competitive on Geometric reasoning, with GPT-5.2 and Qwen3.5-397B slightly
ahead of \ours, suggesting that global extent estimation and holistic spatial
abstraction are not fully solved by object-level alignment alone. By contrast,
\ours retains the best Physical score, indicating that explicit alignment helps
the decoder ground higher-level relational reasoning on cleaner object
evidence.

\noindent\textbf{MMVMBench.}
To evaluate out-of-domain generalization, we further assess \ours{} on the
external benchmark MMVMBench~\cite{zhou2025mmcorr}. Our model achieves an
accuracy of 49.5\% on 1,453 valid test cases using the contrastive retrieval
head, outperforming the native Qwen3-VL-8B baseline by 19.4 points. This gain
shows that the benefits of \ours{} are not limited to in-domain optimization,
but transfer to unseen benchmarks with different data distributions.
Table~\ref{tab:mmvm} provides a detailed breakdown across the eight match types
defined in MMVMBench~\cite{zhou2025mmcorr}. The strongest gains appear on CL,
SZ, RP, OO, and OM, while SP and TM remain closer to the baseline. This pattern
is consistent with the strengths of \ours: it most strongly improves settings
that require stable cross-view association and object-level relational
structure, rather than uniformly boosting every appearance-driven comparison.


\begin{table}[t]
\centering
\caption{Generalization Results on MMVMBench.}
\label{tab:mmvm}
\setlength{\tabcolsep}{3.2pt}
\renewcommand{\arraystretch}{1.06}
\resizebox{\columnwidth}{!}{%
\begin{tabular}{lccccccccc}
\toprule
Method & Overall & CL & SP & TM & SZ & RP & OO & BR & OM \\
\midrule
Qwen3-VL-8B & 30.1 & 27.9 & \best{41.7} & \best{60.0} & 40.3 & 24.0 & 28.0 & 26.6 & 32.7 \\
\rowcolor{ourshade}
\textbf{CrossViewer} & \best{49.5 (+19.4)} & \best{45.8} & 33.3 & 55.3 & \best{71.4} & \best{49.1} & \best{62.4} & \best{33.7} & \best{45.4} \\
\bottomrule
\end{tabular}%
}
\end{table}

\subsection{Ablation Study}

We next analyze which ingredients are responsible for the gains of \ours{} on
\bencheval{}.

\begin{table}[t]
\centering
\caption{Ablation study on \bencheval{}.}
\label{tab:ablation}
\small
\setlength{\tabcolsep}{4pt}
\renewcommand{\arraystretch}{1.06}
\resizebox{\columnwidth}{!}{%
\begin{tabular}{lccccc}
\toprule
Variant & Overall & Corr. & Vis/Occ & Geometric & Physical \\
\midrule
\rowcolor{openshade}
\multicolumn{6}{l}{\textbf{Training Baselines}} \\
Vanilla SFT          & 46.1 & 41.3 & 31.5 & 50.8 & 71.6 \\
w/ box-as-mask       & 51.5 & 68.7 & 50.4 & 41.2 & 61.4 \\
\rowcolor{apishade}
\multicolumn{6}{l}{\textbf{Component Ablations}} \\
w/o \tokenizer{} (mean pool) & 52.7 & 67.0 & 35.0 & 48.8 & 70.0 \\
w/o CrossView Attn   & 47.6 & 54.0 & 35.6 & 45.3 & 72.4 \\
w/o SupCon           & 47.8 & 53.0 & 38.1 & 45.5 & 71.6 \\
\rowcolor{ourshade}
\multicolumn{6}{l}{\textbf{Full Model}} \\
\ours~(full)         & \best{62.7} & \best{83.2} & \best{61.1} & \best{49.1} & \best{74.4} \\
\bottomrule
\end{tabular}%
}
\end{table}

\noindent\textbf{Training Baselines And Mask Quality.}
Table~\ref{tab:ablation} first compares \ours{} against two training baselines.
Vanilla SFT uses the same Qwen3-VL-8B backbone and the same \benchtrain{}
split, but without explicit cross-view alignment, and reaches only 46.1
Overall. This result shows that in-domain instruction tuning alone does not
induce stable cross-view association. Replacing precise masks with box-as-mask
supervision improves performance to 51.5, confirming that explicit object
references are already helpful, but the model still trails the full system by
11.2 points. The remaining gap indicates that coarse localization cannot
preserve the fine boundaries and partial-visibility cues needed for reliable
object-level reasoning across views.

\noindent\textbf{Component Contribution.}
The component ablations isolate the roles of ART, cross-view attention, and
contrastive alignment. Replacing \tokenizer{} with masked-region mean pooling
especially hurts Vis/Occ (61.1 $\rightarrow$ 35.0), showing that a single
pooled descriptor loses local evidence that is crucial for visibility and
occlusion reasoning. Removing CrossView Attn causes the largest overall drop
(62.7 $\rightarrow$ 47.6), confirming that token-level cross-view interaction
is the main mechanism that turns retrieval matches into usable reasoning
evidence. Removing SupCon ($\mathcal{L}_{\mathrm{SC}}$) sharply degrades
Correspondence (83.2 $\rightarrow$ 53.0), indicating that discriminative
cross-view embeddings are essential for stable identity recovery even when the
fusion pathway remains intact.

\begin{table}[t]
\centering
\caption{Hyperparameter sensitivity on \bencheval{}.}
\label{tab:sensitivity}
\small
\setlength{\tabcolsep}{4pt}
\renewcommand{\arraystretch}{1.06}
\resizebox{\columnwidth}{!}{%
\begin{tabular}{lcccc}
\toprule
Factor & Option A & Option B & Option C & Option D \\
\midrule
ART token count & 4: 58.0 & 8: 61.5 & 10: 62.4 & \cellcolor{ourshade}\textbf{16: 62.8} \\
SupCon loss weight & 0.5: 60.0 & \cellcolor{ourshade}\textbf{1.0: 62.4} & -- & -- \\
Triplet loss weight & 0: 62.0 & 0.05: 62.3 & \cellcolor{ourshade}\textbf{0.1: 62.4} & 0.2: 61.6 \\
OCVA depth & \cellcolor{ourshade}\textbf{2: 63.0} & 4: 62.5 & -- & -- \\
\bottomrule
\end{tabular}%
}
\end{table}

\noindent\textbf{Hyperparameter Sensitivity.}
We also vary ART token count, SupCon weight, triplet weight, and OCVA depth
(Table~\ref{tab:sensitivity}). Increasing the ART token count from 4 to 16
steadily improves Overall accuracy from 58.0 to 62.8, with the gain largely
saturating after 10 tokens. This suggests that ART benefits from preserving
enough local structure, but does not require excessively long object token
sequences. SupCon weight 1.0 and triplet weight 0.1 provide the best trade-off
among the tested values, indicating that moderate metric supervision is
sufficient. Finally, a 2-layer OCVA slightly outperforms a 4-layer variant,
suggesting that this benchmark benefits more from precise alignment than from
additional fusion depth.

\noindent\textbf{Retrieval Matching Strategy.}
Table~\ref{tab:match_mode} shows that Greedy and Hungarian perform similarly,
suggesting that the learned similarity space is already sharp enough for simple
matching. In contrast, GT ordering performs substantially worse, indicating a
train-deployment mismatch between oracle ordering and inference-time retrieved
correspondences.

\begin{table}[t]
\centering
\caption{Effect of retrieval matching strategy on \bencheval{}.}
\label{tab:match_mode}
\setlength{\tabcolsep}{4pt}
\renewcommand{\arraystretch}{1.06}
\resizebox{\columnwidth}{!}{%
\begin{tabular}{lccccc}
\toprule
Matching Strategy & Overall & Corr. & Vis/Occ & Geometric & Physical \\
\midrule
\rowcolor{openshade}
\multicolumn{6}{l}{\textbf{Inference-time Matching}} \\
\rowcolor{ourshade}
Greedy                      & \best{62.7} & \best{83.2} & \best{61.1} & 49.1        & \best{74.4} \\
Hungarian                   & 61.9        & 82.6        & 60.6        & \best{49.5} & 73.9        \\
\rowcolor{apishade}
\multicolumn{6}{l}{\textbf{Analysis-only Reference}} \\
GT ordering                 & 52.8        & 69.4        & 51.3        & 42.2        & 69.3        \\
\bottomrule
\end{tabular}%
}
\end{table}

\FloatBarrier
\subsection{Qualitative Analysis}

\begin{figure}[!t]
  \centering
  \includegraphics[width=\columnwidth]{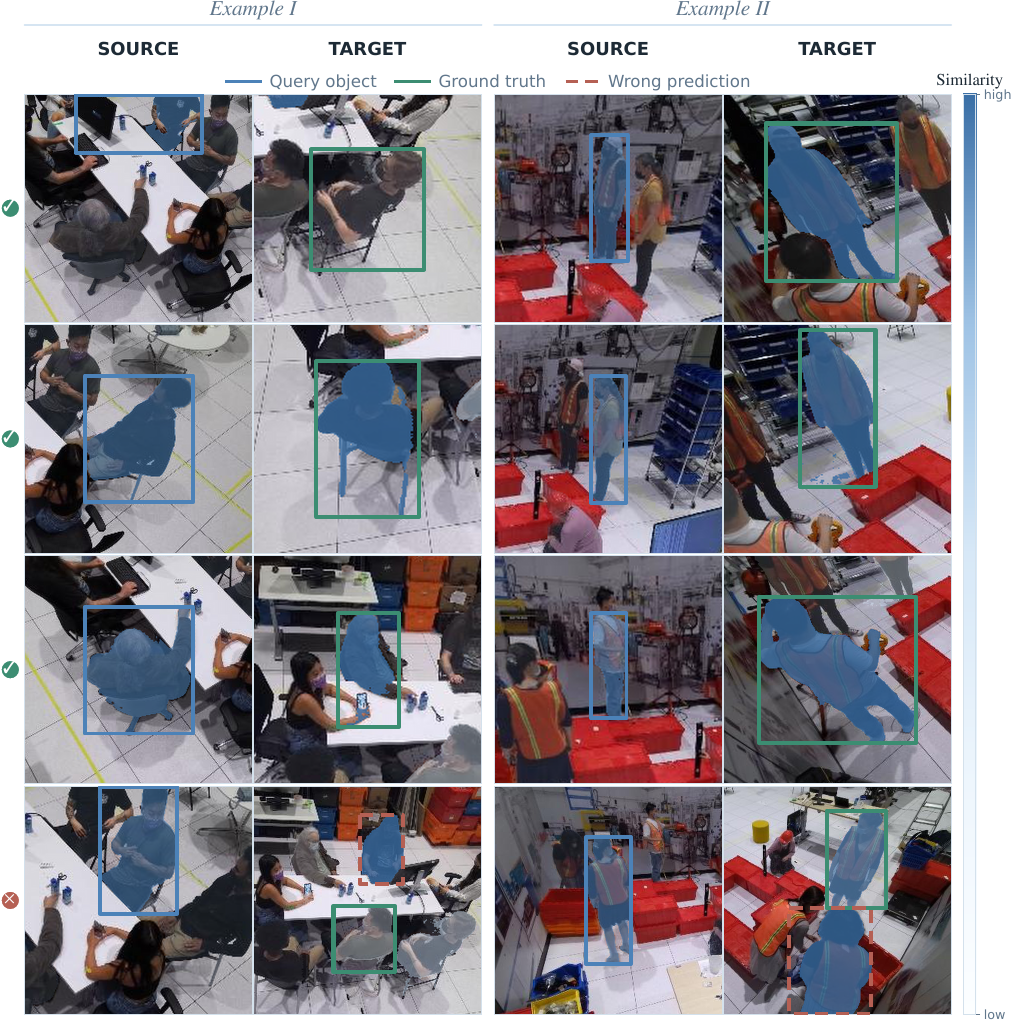}
  \caption{Qualitative cross-view retrieval results.}
  \Description{Two example groups of cross-view retrieval results, with source queries on the left and similarity-shaded target candidates on the right.}
  \label{fig:crossattn_qual}
\end{figure}

We next visualize the learned correspondence space and summarize the source and
task composition of \benchtrain{} and \bencheval{}. Fig.~\ref{fig:small_panels_top} (c) embeddings from the two views are strongly intermixed before training and yield
only 20.5\% top-1 matching on the shown Q1 subset; after training, matched
objects form tighter neighborhoods and top-1 matching rises to 52.7\%,
supporting the role of contrastive supervision. As shown in
Fig.~\ref{fig:crossattn_qual}, most queries recover the
correct target despite viewpoint changes, while the failure case highlights the
remaining ambiguity between nearby instances.

\section{Conclusion}
\label{sec:conclusion}

We introduced \suite{}, which combines \benchtrain{}, \bencheval{}, and \ours{} for multi-view
spatial intelligence. On \bencheval{}, \ours{} reaches 62.7\% overall accuracy,
outperforming Qwen3-VL-8B by 20.0\%. Our analyses show that
\tokenizer{}, contrastive alignment, and cross-view attention are all
important. We hope \bencheval{} will support future work on richer multi-view
settings and open-ended spatial reasoning in MLLMs.

\begin{acks}
We thank the authors of Ego-Exo4D, EgoHumans, MessyTable, and MMPTracking
for making their datasets publicly available.
\end{acks}

\clearpage
\bibliographystyle{ACM-Reference-Format}
\bibliography{refs}

\end{document}